\documentclass[10pt,twocolumn,letterpaper]{article}

\usepackage[pagenumbers]{arxiv}

\usepackage{graphicx}
\usepackage{amsmath}
\usepackage{amssymb}
\usepackage{booktabs}
\usepackage{pifont}
\usepackage{subcaption}
\usepackage{booktabs}
\usepackage{caption}
\usepackage{float}
\usepackage{multirow}

\usepackage[pagebackref,breaklinks,colorlinks]{hyperref}

\usepackage[capitalize]{cleveref}
\crefname{section}{Sec.}{Secs.}
\Crefname{section}{Section}{Sections}
\Crefname{table}{Table}{Tables}
\crefname{table}{Tab.}{Tabs.}

\newcommand{\cihppair}[0]{CIHP$_{\textrm{P}}$\xspace}
\newcommand{\csppair}[0]{CSP$_{\textrm{P}}$\xspace}
\newcommand{\cihpmulti}[0]{CIHP$_{\textrm{M}}$\xspace}
\newcommand{\cspmulti}[0]{CSP$_{\textrm{M}}$\xspace}
\newcommand{\cihppairinst}[0]{CIHP$_{\textrm{P}}^{\textrm{-inst}}$\xspace}
\newcommand{\csppairinst}[0]{CSP$_{\textrm{P}}^{\textrm{-inst}}$\xspace}
\newcommand{\cocosemseg}[0]{COCO$^{\textrm{-sem}}$\xspace}
\newcommand{\cssemseg}[0]{CS$^{\textrm{-sem}}$\xspace}

\newcommand{\ourmethodname}[0]{RESI\xspace}

\newcommand{\panoinffinal}[0]{ESF-OMI\xspace}
\newcommand{\cmark}{\ding{51}}%
\newcommand{\xmark}{\ding{55}}%

\newcommand*{\affmark}[1][*]{\textsuperscript{#1}}
\DeclareRobustCommand{\dgrponetxt}[0]{{\texttt{D1}}\xspace}
\DeclareRobustCommand{\dgrptwotxt}[0]{{\texttt{D2}}\xspace}
\DeclareRobustCommand{\dgrpthreetxt}[0]{{\texttt{D3}}\xspace}
\DeclareRobustCommand{\supcihpp}[0]{{\texttt{\cihppair}}\xspace}
\DeclareRobustCommand{\supcihpm}[0]{{\texttt{\cihpmulti}}\xspace}
\DeclareRobustCommand{\supcspp}[0]{{\texttt{\csppair}}\xspace}
\DeclareRobustCommand{\supcspm}[0]{{\texttt{\cspmulti}}\xspace}

\begin{document}

\title{Resolving Inconsistent Semantics in Multi-Dataset Image Segmentation}

\author{
    Qilong Zhangli\affmark[1]\quad 
    Di Liu\affmark[1]\quad 
    Abhishek Aich\affmark[2]\quad 
    Dimitris N. Metaxas\affmark[1]\quad 
    Samuel Schulter\affmark[2]\quad \\
    {
        \affmark[1]Rutgers University\quad
        \affmark[2]NEC Laboratories America
    }
}

\maketitle

\begin{abstract}
  Leveraging multiple training datasets to scale up image segmentation models enhances robustness and semantic understanding. Individual datasets have well-defined ground truth with non-overlapping mask layouts and mutually exclusive semantics. However, merging them for multi-dataset training disrupts this harmony and leads to semantic inconsistencies. For instance, the class ``person'' in one dataset and the class ``face'' in another will require multilabel handling for certain pixels. Existing methods struggle with this setting, particularly when evaluated on label spaces mixed from the individual training sets. To address these challenges, we introduce a simple yet effective multi-dataset training approach by integrating language-based embeddings of class names and label space-specific query embeddings. Our method maintains high performance regardless of the underlying inconsistencies between training datasets. Notably, on four benchmark datasets with label space inconsistencies during inference, we outperform previous methods by 1.6\% mIoU for semantic segmentation, 9.1\% PQ for panoptic segmentation, 12.1\% AP for instance segmentation, and 3.0\% in the newly proposed PIQ metric.
\end{abstract}
\section{Introduction}
\label{sec:intro}
The advancement of image segmentation hinges significantly on scaling models to improve robustness and deepen semantic understanding~\cite{cheng2021mask2former,kirillov_cvpr_19_panoseg}. This scaling necessitates an extensive collection of annotated datasets~\cite{lin_eccv_14_mscoco,neuhold_iccv_2017_vistas,zhou_cvpr_2017_ade20k}. However, creating such datasets is both costly and labor-intensive~\cite{Cordts2016Cityscapes}. Models like SAM~\cite{kirillov2023segment, ravi2024sam} and HQ-SAM~\cite{ke2024segment} have demonstrated remarkable capabilities with meticulously curated datasets, but these are extremely expensive to produce and often lack comprehensive semantic labels. An alternative strategy involves leveraging existing datasets that are already annotated. Individually, these datasets maintain a consistent label space, but when combined, their labels may conflict, introducing challenges in maintaining semantic consistency across the datasets.

\begin{figure*}[t]\centering
  \includegraphics[width=1.0\linewidth]{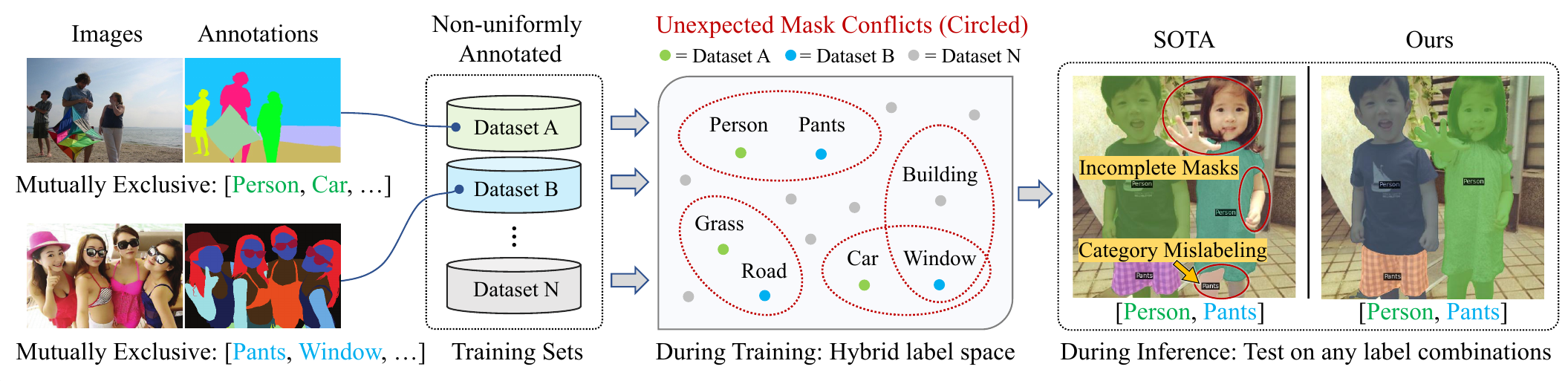}
  \vspace{-2em}
  \caption{Leveraging multiple datasets for training segmentation models increases robustness and semantic understanding. However, existing methods (1) fail to capture the full masks of objects, such as ``person'' category, (2) often predict incorrect labels, for instance, mistaking ``legs'' for ``pants''. This issue is caused by unexpected conflicts in multiple label spaces, although each dataset (\textit{A} and \textit{B}) has consistent ground truth mask layouts that provide non-overlapping and mutually-exclusive semantics.
  }
  \vspace{-1em}
  \label{fig:teaser}
\end{figure*}

The concept of multi-dataset training, although progressing in various domains such as object detection and semantic segmentation~\cite{zhao_eccv20_multidatasetdet,zhou_cvpr22_simple,bevandic_wacv22_multidomain_semseg}, presents unique challenges when applied to more complex segmentation scenarios where combining datasets leads to inconsistent semantics. As the number of labels increases, traditional assumptions, such as exclusive per-pixel labeling, become less practical. For example, when combining two datasets, the individual semantics may violate the mutual exclusivity assumption, such as with ``person'' and ``clothing'' (retail), ``road'' and ``lane marking'' (mobility), or ``person'' and ``face'' (surveillance), as shown in Fig.~\ref{fig:teaser}.

We found that even state-of-the-art base models, like Mask2Former (M2F)~\cite{cheng2021mask2former}, combined with existing multi-dataset training strategies~\cite{zhao_eccv20_multidatasetdet,zhou_cvpr22_simple} falter in this setting. While M2F is equipped to meet some of our requirements, it falls short in dealing with the intricacies presented by multi-dataset training (see Fig.~\ref{fig:teaser}). This challenge underscores the necessity of not only a more adaptable segmentation model but also a revised approach to benchmarking and ground truth annotation~\cite{zhao_eccv20_multidatasetdet,zhou_cvpr22_simple}.

In response to these challenges, we propose a novel multi-dataset training framework, \textbf{\ourmethodname} (\textbf{Res}olving \textbf{I}nconsistent Semantics in Multi-Dataset Image Segmentation), specifically designed to address the issue of inconsistent semantics in label spaces when training across multiple datasets. Our approach extends the baseline Mask2Former framework~\cite{cheng2021mask2former} with the following key modifications: 

\textit{First}, we replace the fixed-size label space classifier with vision \& language embeddings from CLIP~\cite{radford_arxiv_2021}, similar to works like~\cite{li_iclr_22}. This serves two purposes: \textit{(1)} mapping all categories into a single, consistent space that preserves semantic relations from the pre-trained vision-language model, and \textit{(2)} enabling our model to operate with any combination of training set labels at inference. 

\textit{Second}, we introduce label space-specific query embeddings added as residuals to the transformer decoder in Mask2Former. These learnable embeddings condition the decoder, and thus mask predictions, on the label spaces by retrieving the relevant query embeddings. These enhancements equip our model to effectively adapt to and reconcile the inconsistencies inherent in multi-dataset training.

To validate our proposed framework \ourmethodname, we conduct a series of experiments where we train on various groups of multiple datasets. We compare \ourmethodname with multiple baselines on two newly created benchmarks that specifically evaluate situations where the test-time label space is a combination of the individual training datasets.
It is important to note that no single segmentation task - semantic, instance or panoptic - adequately benchmarks scenarios that include (instance-aware) ``thing'' and ``stuff'' classes and that allows semantic overlaps (multilabel). Hence, we evaluate on all three tasks, as well as a newly introduced metric. The Panoptic Instance Quality (PIQ) is innovatively designed to combine the per-pixel classification strength of panoptic segmentation with the ability to accommodate overlapping masks inherent in instance segmentation, thus providing a more comprehensive assessment of segmentation models in mixed label space scenarios.
Averaged over all benchmarks, \ourmethodname outperforms the best baseline by 1.6\% mIoU for semantic segmentation, 12.1\% AP for instance segmentation, 9.1\% PQ for panoptic segmentation, and 3.0\% on the newly introduced PIQ metric, demonstrating its ability to handle semantic conflicts and overlapping masks. We also demonstrate on-par or better performance on standard multi-dataset benchmarks that evaluate models on the individual label spaces of the training datasets.

\section{Related Work}
\label{sec:related_work}

\paragraph{Image Segmentation:} Understanding and interpreting visual data is a core challenge in computer vision encompassing various tasks~\cite{cheng2024yolo,zhangli2024layout,xia2022sign,sayadi2022harnessing,liu2024lepard,wen2024second}, with segmentation being one of the most important. Different formulations have been proposed, including semantic segmentation~\cite{deeplabv3plus2018,long_cvpr_15_fcn,yu_iclr_16_dilated,zhao_cvpr_17_pspnet,liu2022transfusion,chang2022deeprecon,he2023dealing,gao2024training} (pixels are assigned a semantic class without distinction of instances from the same class), instance segmentation~\cite{li_arxiv_18_tascnet,li_cvpr_19_attentionguided_pano,mohan2021ijcv_efficientps,porzi_cvpr_19_seamseg,xiong_cvpr_19_upsnet,athar2023tarvis, zhangli2022region,gu2024dataseg} (separates instances, but does not consider ``stuff'' categories - amorphous non-countable objects like sky or road), and panoptic segmentation~\cite{kirillov_cvpr_19_panoseg,panoptic_deeplab_2020,kirillov_cvpr_19_panofpn,li_arxiv_18_tascnet,li_cvpr_19_attentionguided_pano,mohan2021ijcv_efficientps, porzi_cvpr_19_seamseg,xiong_cvpr_19_upsnet,athar2023tarvis,sun2024remax,kappeler2024few,chen2023generalist} (handles all categories and separates instances). With the goal of universal and robust segmentation, the latest research focused on building unified architectures to handle all three task formulations simultaneously. Building on Transformer architectures \cite{vaswani_neurips_17_attention,carion_eccv_20_detr,max_deeplab_2021}, MaskFormer~\cite{cheng2021maskformer}, Mask2Former~\cite{cheng2021mask2former} and UniFormer~\cite{li2023uniformer} are good examples of such unified architectures. Our work extends the Mask2Former architecture to better handle semantic inconsistencies when training from multiple datasets, which is a key part when scaling segmentation models.

\vspace{-0.4em}
\paragraph{Scaling Data for Segmentation Models:} With the same motivation of universal segmentation comes the requirement of training from large-scale data in order to increase model robustness and semantic understanding. The Segment Anything project~\cite{kirillov2023segment, ravi2024sam} demonstrated possibilities but required significant annotation effort and does not address semantic inconsistencies. On the other hand, more cost effective solutions are proposed with open-vocabulary and multi-dataset training.
The goal of \emph{open-vocabulary segmentation} is to extend semantic understanding to unseen categories without explicit mask annotations. Building on recent advances in vision \& language models~\cite{Chen_ECCV_20_UNITER,jia_icml_21_align,ALBEF, Lu_neurips_19_ViLBERT,radford_arxiv_2021,Sun_iccv_19_VideoBERT,khan_eccv22_simla}, open-vocabulary variants for semantic segmentation~\cite{li_iclr_22,rao2021denseclip,xu2021simple,zhou2021denseclip} and panoptic~\cite{xu2022odise,zhang2023simple,ding2022open,zou2022xdecoder} segmentation have been proposed. While such works can be trained from multiple datasets, the typical settings and benchmarks for open-vocabulary segmentation do not explicitly challenge the model with inconsistent semantics.
The goal of \emph{multi-dataset training} is to leverage existing datasets with various semantic annotations to improve generalization robustness. Methods have been introduced for object detection~\cite{zhao_eccv20_multidatasetdet, rame_arxiv18_omnia,yao_arxiv20_crossdataset_objdet,zhao_arxiv20_catextended_objdet, zhou_cvpr22_simple}, semantic segmentation~\cite{MSeg_2020_CVPR,bevandic_wacv22_multidomain_semseg,kemnitz_mlmi_18_multidataset_seg_medical,kim_eccv22_multiseg}, and panoptic segmentation~\cite{zhou2023lmseg}.
However, none of the works in open-vocabulary or multi-dataset segmentation investigate or evaluate semantic inconsistencies in label space that arise when combining multiple datasets. In this work, we highlight this issue, demonstrate limitations of existing methods, extend Mask2Former~\cite{cheng2021mask2former} to handle such inconsistencies, and propose methods for effective evaluation.

\section{Method} 
\label{sec:method}

\subsection{Motivation}
One efficient way to scale up a segmentation model is through multi-dataset training, which involves training a single model to perform accurately across various datasets. However, as we increase the number of classes in such training, we often face label space inconsistencies. Our goal is to facilitate multi-dataset training while accommodating these inconsistencies. A straightforward approach, similar to that in~\cite{zhou2023lmseg}, would be to condition the decoder of an existing MaskFormer~\cite{cheng2021maskformer} model with the label space. However, our early experiments showed that this method did not significantly improve upon the baseline MaskFormer model. (Tab. \ref{tbl:results_3tables_merged} and Fig. \ref{fig:qualitative_results}) We believe this is due to two primary reasons: Firstly, conditioning the MaskFormer decoder on a per-dataset label space restricts the model from understanding novel combinations of categories from different datasets. During inference, the model struggles with new label spaces it did not encounter during training. Secondly, this approach does not adequately address the issue of inconsistent annotations in the training data. For instance, when the mask of ``person'' overlaps with the mask of ``pants'', the model's confidence in these regions diminishes, resulting in ambiguous predictions (see Fig. \ref{fig:teaser}). 

\subsection{Preliminaries: Model Framework}
\label{sec:method_mask2former}

Given an image $I$, our segmentation model is designed to predict multiple masks, potentially overlapping, with each mask being associated with a semantic category $c \in \{1, \ldots, C\}$. This approach deviates from the conventional semantic or panoptic segmentation settings where typically only one label per pixel is predicted with no overlaps allowed.

The set of $C$ categories annotated in each dataset is divided into instance-aware ``thing'' categories (countable objects like cars or persons) and ``stuff'' categories (amorphous, non-countable objects like sky or road). For ``stuff'' categories, instances are irrelevant and thus multiple masks of the same ``stuff'' category are merged.

Our work builds on Mask2Former (M2F)\cite{cheng2021mask2former,cheng2021maskformer} but can be easily integrated into other models. This model processes the input image $I$ with a combination of a standard visual backbone (CNN or Transformer) and a Transformer-based encoder, which outputs multi-scale visual features. Then, a Transformer-based decoder predicts a set of $N$ masks $m_i \in [0,1]^{H \times W}$, with $i\in{1, \ldots, N}$, along with class probabilities $p_i \in \mathbb{R}^{C+1}$, where $H,W$ are downscaled image dimensions and $C+1$ is the number of categories including background.

\begin{figure*}[t]\centering
    \includegraphics[width=1.0\linewidth]{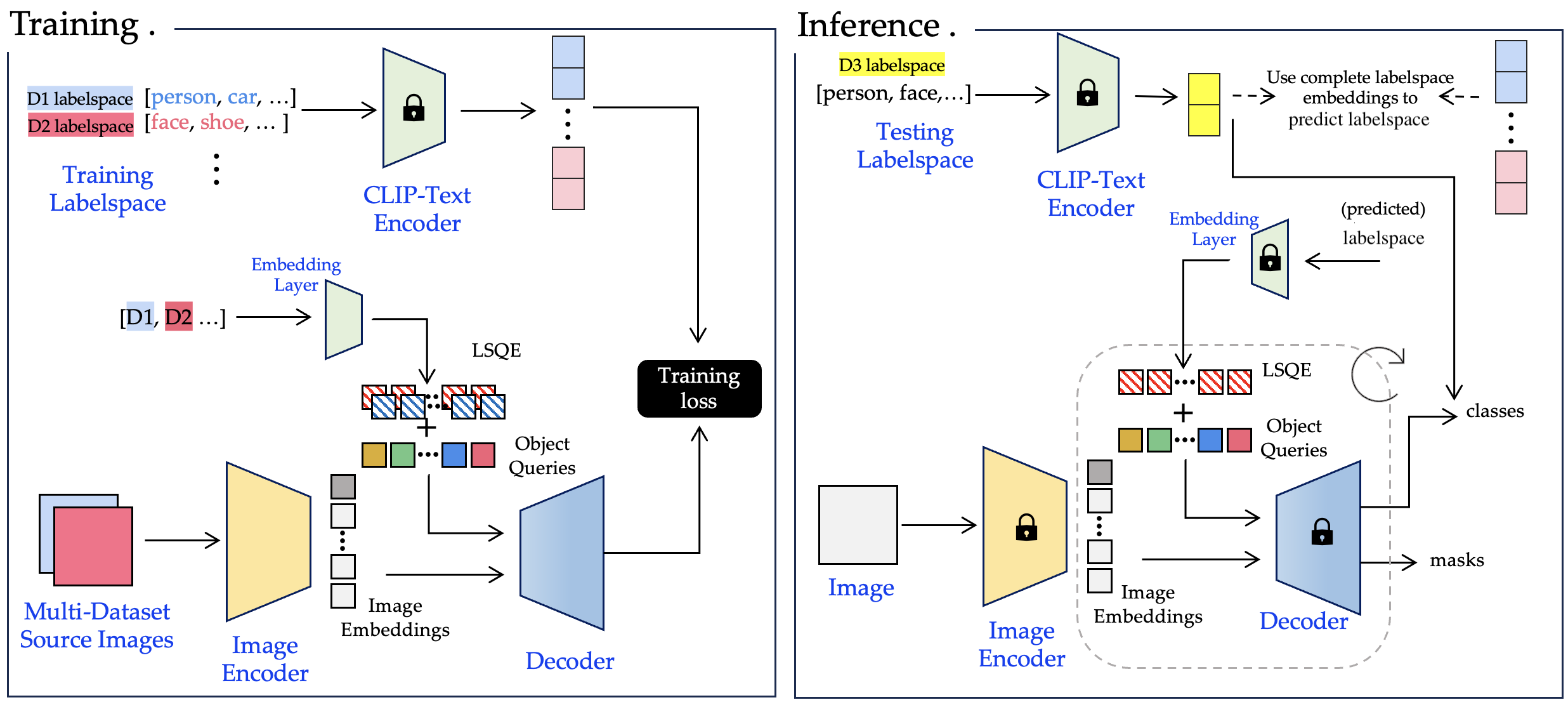}
    \caption{
    \textbf{Overview of our proposed framework}: \textit{Training}. We build upon Mask2Former~\cite{cheng2021mask2former} and replace the fixed-label space classifier with language-based embeddings. We introduce learnable label space-specific query embeddings (LSQE) that are added to the decoder in order to handle conflicting label spaces that arise in the multi-dataset setting.
    \textit{Inference}. Given a new label space for inference -- any combination of the categories of the training datasets -- the model first predicts what training label spaces can ``serve'' the test-label space by matching the text-embeddings of the class names. This process selects the LSQEs that are needed for inference. Then, the decoder of the model runs for each selected LSQE -- at most $K$, the number of training datasets.}
    \vspace{-1em}
    \label{fig:method_overview}
\end{figure*}

Note that this model formulation naturally handles both ``thing'' and ``stuff'' categories and theoretically also allows for mask overlaps. The decoder is a multi-layer Transformer that takes $N$ learnable embeddings $e^O_i$ (or $O$bject queries) as input, and that performs self-attention among the $N$ object queries as well as cross-attention with the image features in each layer. The high-level architecture is evident in Fig.~\ref{fig:method_overview}. The objective of M2F can be defined as

\begin{align}
    \mathcal{L} = \sum_{i=1}^{N} l_{\mathrm{C}}\left(p_i, p^{\mathrm{*}}\right) + \left[ p^{\mathrm{*}} \neq \emptyset \right] \, l_{\mathrm{BC}}\left(m_i, m^{\mathrm{*}}\right) \;,
    \label{eq:m2f_loss}
\end{align}

where $l_{\mathrm{(B)C}}$ stands for (binary) cross-entropy loss, $[\cdot]$ is the indicator function, and $\{p,m\}^{\mathrm{*}}$ indicate ground truth category ($p$) and mask ($m$). To compute the loss function, a bipartite matching algorithm is employed to optimally pair predictions with ground truth. Once the matching is established, the loss is calculated based on these pairs. Please refer to~\cite{cheng2021mask2former,cheng2021maskformer} for more details.

Next, we outline the key adaptations for the multi-dataset training setting: language-based classifiers (Sec.~\ref{sec:method_multi_dataset_training}) and label space-specific query embeddings (Sec.~\ref{sec:method_partwhole_relations}).

\subsection{Language Embeddings as Classifiers}
\label{sec:method_multi_dataset_training}
To train from multiple datasets, we need to handle heterogeneous label spaces.  While some prior works resolve the conflicting label spaces manually~\cite{zhao_eccv20_multidatasetdet} or via post-training optimization~\cite{zhou_cvpr22_simple}, we use language embeddings from the CLIP~\cite{radford_arxiv_2021} text encoder as a simple but effective solution. Instead of directly predicting a probability distribution $p_i$ with a fixed-label space classifier, our model predicts an embedding vector $e^I_i \in \mathbb{R}^{d}$ for each object query $i \in \{1, \ldots, N\}$. We then use CLIP's (pre-trained and frozen) text-encoder to compute embedding vectors $e^T_c$ for each category $c$. Based on these two embedding vectors, we can define the class probability $p_i$ as

\begin{equation}
    p_i = \mathrm{S}\left( \frac{1}{\tau} \left[ \langle e^I_i, e^T_1 \rangle, \ldots, \langle e^I_i, e^T_C \rangle, \langle e^I_i, e^T_\emptyset \rangle \right] \right) \;,
    \label{eq:vl_prob}
\end{equation}

where $\mathrm{S}(\cdot)$ is the SoftMax function, $\langle\cdot,\cdot\rangle$ denotes the dot product and $e^T_\emptyset$ is an all-zero vector representing the ``no-object'' class, following \cite{gao2021open,zareian_cvpr_21}.  We set the temperature $\tau$ to $0.01$~\cite{radford_arxiv_2021}. All embedding vectors $e^{\{I,T\}}$ are $\ell_2$-normalized.  The class probability $p_i$ can be plugged into Eq.~\ref{eq:m2f_loss} for training.  During training, we first sample a dataset $k \in \{1, \ldots, K\}$, which defines the active label space $\mathcal{L}_k$ that is used in the current iteration. The pre-defined embedding space of the vision-and-language model naturally handles the different label spaces. While each category has its own spots in the embedding space, different names of the same semantic category (e.g., synonyms like ``sofa'' and ``couch'') will be close due to the large-scale pre-training of CLIP~\cite{radford_arxiv_2021}.

\subsection{Label-space Specific Query Embeddings}
\label{sec:method_partwhole_relations}

While \textbf{language-based classifiers like CLIP} are powerful in mapping diverse label spaces into a unified embedding space, they alone \textbf{are not sufficient to fully resolve inconsistent semantics across multiple datasets}. CLIP helps align categories with similar meanings but does not address overlapping masks, hierarchical inconsistencies (e.g., ``person'' vs. ``sunglasses on a person''), or conflicting annotations in multi-dataset training.

\paragraph{Inconsistent Semantics.} In multi-dataset training, semantic conflicts can naturally emerge when different datasets annotate objects at varying levels within a semantic hierarchy. An example of such a hierarchy is the \textit{part-whole} relationship. For instance, Dataset \textit{A} might annotate ``car'', while Dataset \textit{B} focuses on parts, like ``wheel''. Inconsistencies can also arise between datasets. For example, Dataset \textit{C} could annotate ``person'' in natural images, whereas Dataset \textit{D} might annotate accessories like ``sunglasses'' in product photos. Individually, each dataset maintains semantic consistency and follows the principle that one pixel should belong to exactly one semantic class and one object instance (if it belongs to a ``thing'' class). However, inconsistencies emerge when these label spaces are combined, resulting in scenarios where one pixel might correspond to two different semantic classes and instance masks.

\paragraph{Naive approach.} 
\label{para:naive_approach}

While instance segmentation methods naturally handle overlaps, they do not integrate ``stuff'' categories. However, the Mask2Former framework, as described above, provides this flexibility, offering state-of-the-art instance segmentation results while handling both ``thing'' and ``stuff'' categories. We also experimented with Mask2Former in this setting. However, even after incorporating language-based embeddings as classifiers, the resulting models often struggled with semantically inconsistent relationships between label spaces (details provided in the supplementary materials).

\paragraph{Our solution.} To resolve potential conflicts due to inconsistent semantic relations when training from multiple datasets, we introduce label space-specific query embeddings (\textbf{LSQE}) in the decoder-transformer of Mask2Former. These are $K$ learnable embeddings $e^L_k$ (same dimension size as $e^O_i$), one for each of the $K$ training datasets. When training from an image of dataset $k$, we add the corresponding LSQE to each of the $N$ object query embeddings, obtaining new inputs to the decoder as $e^{O,k}_i = e^O_i + e^L_k$. Hence, LSQEs introduce a decomposition of object queries into object-specific and label space-specific parts, which we illustrate in Fig.~\ref{fig:method_overview}.

\paragraph{Underlying intuition.} LSQEs give the model the freedom to internally resolve potential conflicts, while at the same time leverage common information from multiple datasets. Similar to UniDet~\cite{zhou_cvpr22_simple}, the label space-specific information is multiplexed through the network. Images from multiple datasets and label spaces go into the network and predictions for individual datasets are made. While \cite{zhou_cvpr22_simple} computes per-label space probabilities with only the last classification layer, our LSQEs allow the model to consider this information throughout the whole decoder stage, which can also influence the mask predictions.

\subsection{Inference with LSQEs}
\label{sec:method_inference}

Associated with each of the $K$ LSQEs are the class embedding vectors for each individual dataset $\{ e^{T}_{c,k} \}$, where $k$ indices the label-space and $c$ the class within that label-space. During inference, our model can take any combination of categories from all training datasets as the input label space, which are also encoded with the CLIP text-encoder to $\{ e^{T}_{c,*} \}$, where $*$ refers to the inference-time label space. 

To do inference with LSQEs, we first find the closest match between each test class embedding $e^{T}_{c,*}$ and the training class embeddings $\{ e^T_{c,k} \}$. Each match reveals a label-space specific index $k$, which we collect to the set $D = \{ k \}$. This set can have at most $K$ elements. If any test-time class $e^{T}_{c,*}$ has equal similarity to multiple classes from different datasets, we include both label-space specific indices. This happens when multiple datasets share the same category, like ``person''. We then run the decoder $|D|$ times with the corresponding LSQEs $e^L_k$ and collect the $N \cdot |D|$ predictions consisting of masks and embedding vectors $e^I_i$. This strategy is highly effective in preserving label-space consistency and accuracy but can increase the inference time by a small factor (see Tab.~\ref{table:time_rebuttal}). The per-prediction classification is computed as the dot product between $e^I_i$ and all test-time class embeddings $e^{T}_{c,*}$.  Fig.~\ref{fig:method_overview} illustrates the inference process.

\section{Experiments}
\label{sec:experiments}

\subsection{Experimental Settings}

\paragraph{Training datasets.} Since our models all train from multiple datasets, we define a list of three different datasets combinations (details provided in the supplementary materials):
\begin{itemize}
    \vspace{-0.3em}
    \item \dgrponetxt: COCO~\cite{lin_eccv_14_mscoco}, ADE20K~\cite{zhou_cvpr_2017_ade20k}, and Vistas~\cite{neuhold_iccv_2017_vistas}
    \vspace{-0.6em}
    \item \dgrptwotxt: COCO~\cite{lin_eccv_14_mscoco} and  CIHP~\cite{gong_eccv18_cihp}
    \vspace{-0.6em}
    \item \dgrpthreetxt: CityScapes~\cite{Cordts2016Cityscapes} and CityScapesParts~\cite{geus_cvpr21_panopticparts}
    \vspace{-0.3em}
\end{itemize}
The datasets COCO~\cite{lin_eccv_14_mscoco}, ADE20K~\cite{zhou_cvpr_2017_ade20k}, Vistas~\cite{neuhold_iccv_2017_vistas}, and CityScapes~\cite{Cordts2016Cityscapes} are standard benchmarks for panoptic segmentation, defining label spaces of sizes 133, 150, 48, and 19, respectively. CIHP~\cite{gong_eccv18_cihp} annotates human parts such as faces, hair, arms, and clothing, but not the entire person as a whole. Similarly, CityScapesParts (CSP)~\cite{geus_cvpr21_panopticparts} annotates parts of humans and cars (e.g., license plates, windshields), but not the complete objects. The goal for groups \dgrptwotxt and \dgrpthreetxt is for the models to combine whole-object annotations from one dataset with part annotations from another dataset.

\begin{figure}[!htbp]
    \centering
    \raggedleft
    \includegraphics[width=\linewidth]{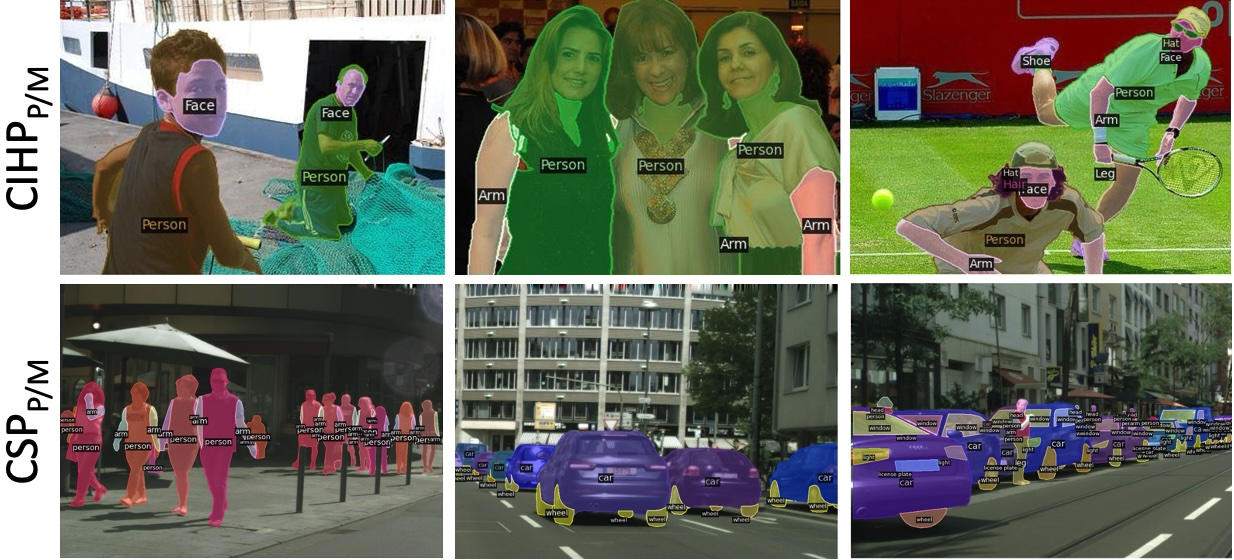}
    \caption{
    Examples of the mixed-label space evaluation-only datasets. Each row shows two examples from CIHP/CSP$_{\textrm{P}}$ (left and middle) and one from CIHP/CSP$_{\textrm{M}}$ (right).}
    \label{fig:cihp_csp_pair_multi}
    \vspace{-1em}
\end{figure}

\paragraph{Benchmarking Image Segmentation.\label{sec:benchmarkingSeg}} Effectively scaling segmentation requires a critical evaluation of current benchmarking standards for semantic, instance, and panoptic segmentation. While these tasks offer a solid evaluation framework, they fall short when dealing with complex scenarios involving overlapping categories. 
Scaling segmentation increases the semantic label space, which includes ``thing'' and ``stuff'' classes and necessitates multi-label assignments for pixels where semantics overlap, e.g., ``person'' and ``face''. Neither semantic segmentation (not instance-aware and no overlaps), instance segmentation (no ``stuff'' categories), or panoptic segmentation (no overlaps) provide a comprehensive benchmark.
Thus, in addition to traditional evaluation metrics for semantic, panoptic, and instance segmentation, we propose a novel approach to benchmarking called Panoptic Instance Quality (PIQ).
It combines the advantages of per-pixel classification from panoptic segmentation and the allowance for overlapping masks in instance segmentation to better measure a model's accuracy in a unified, scalable manner. We achieve this by averaging the Average Precision (AP) score of all instances in the ``thing'' categories and the Panoptic Quality (PQ) score for uncountable background categories (``stuff''). We also include the variations of PIQ like PIQ50 and PIQ75 which follow a similar convention as used in other metrics like AP (Average Precision) in object detection, where AP50 and AP75 represent the precision score calculated at different Intersection over Union (IoU) thresholds.

\paragraph{Mixed label space benchmarks.}To properly evaluate the ability of multi-dataset models to handle any combination of label spaces $A$ and $B$ of the individual datasets, we build a dataset with label space $C$ that has the following property. Label space $C$ must contain two partitions that contain categories that come exclusively from either training dataset, i.e., $| C \cap A \setminus B | > \emptyset \land |C \cap B \setminus A| > \emptyset$. (This can be easily extended to more than two training datasets.) To build such a dataset, we can use CIHP~\cite{gong_eccv18_cihp} and CSP~\cite{geus_cvpr21_panopticparts} that annotate parts of persons and cars. In both cases, putting all masks together results in the super-category, person or car, which is not part of CIHP or CSP themselves, but is part of COCO~\cite{lin_eccv_14_mscoco} and CityScapes~\cite{Cordts2016Cityscapes}. We do not use all parts, because otherwise the super-categories are fully covered by parts. Hence, we create multiple separate datasets by combining subsets of parts with the super-category. First, we use all parts individually which defines the datasets CIHP$_{\textrm{pair}}^i$ and CSP$_{\textrm{pair}}^i$, which all contain two labels. Furthermore we define 4 datasets that use multiple parts together with the super-category, CIHP$_{\textrm{multi}}^i$ and CSP$_{\textrm{multi}}^i$. Fig.~\ref{fig:cihp_csp_pair_multi} shows some examples and details are in the supplementary materials. To report results, we evaluate models on all individual datasets and then average the panoptic quality (PQ)~\cite{kirillov_cvpr_19_panoseg} to form four benchmarks: \cihppair, \cihpmulti, \csppair, \cspmulti (more benchmark details are provided in the supplementary materials).

\paragraph{Evaluation Settings.}
We do two types of quantitative evaluations. Firstly, we conduct a \emph{mixed-label space evaluation} utilizing our newly created dataset annotations that blend classes from multiple datasets, including \cihppair, \cihpmulti, \csppair, and \cspmulti. Secondly, we perform a \emph{per-label space evaluation}, following the approach of prior work~\cite{zhou2023lmseg}. In this approach, we assess the model's performance on each of the individual datasets within our defined dataset groups (\dgrponetxt, \dgrptwotxt, \dgrpthreetxt).

\paragraph{Metrics.}
As discussed before, no single task - semantic, instance or panoptic - provides a comprehensive benchmark when semantic inconsistencies exist. Hence, for quantitative analysis, we employ metrics from all tasks: Intersection over Union (IoU), Average Precision (AP), and Panoptic Quality (PQ). Moreover, we add our newly proposed Panoptic Instance Quality (PIQ). Each metric provides unique insights into the model's performance. Notably, PIQ is instrumental in evaluating performance in the proposed panoptic-instance segmentation.

\paragraph{Baselines.}We compare our method against three relevant baselines. (1) LMSeg~\cite{zhou2023lmseg} is a recently proposed state-of-the-art model for multi-dataset image segmentation (and the only prior work on panoptic segmentation, to the best of our knowledge)\footnote{To report results for LMSeg~\cite{zhou2023lmseg}, we train the model ourselves with the latest Mask2Former~\cite{cheng2021mask2former} framework, which gives higher PQ values than reported in~\cite{zhou2023lmseg} and is a fairer comparison with equal training settings.}. (2) We extend the idea of a dataset-specific classification head from UniDet~\cite{zhou_cvpr22_simple}, which was developed for object detection, to segmentation. (3) Mask2Former~\cite{cheng2021mask2former} with language-embeddings as classifier, see Sec.~\ref{sec:method_multi_dataset_training}. This baseline is the same as our \ourmethodname model, but without the label space-specific query embeddings.

\paragraph{Model training.} For detailed information on model training, hyperparameters, and the panoptic inference algorithm, please refer to the supplementary materials.

\subsection{Evaluation on Mixed Label Spaces}
\label{sec:4_2}
\begin{figure*}[!htbp]\centering
    \vspace{-5pt}
    \includegraphics[width=\textwidth]{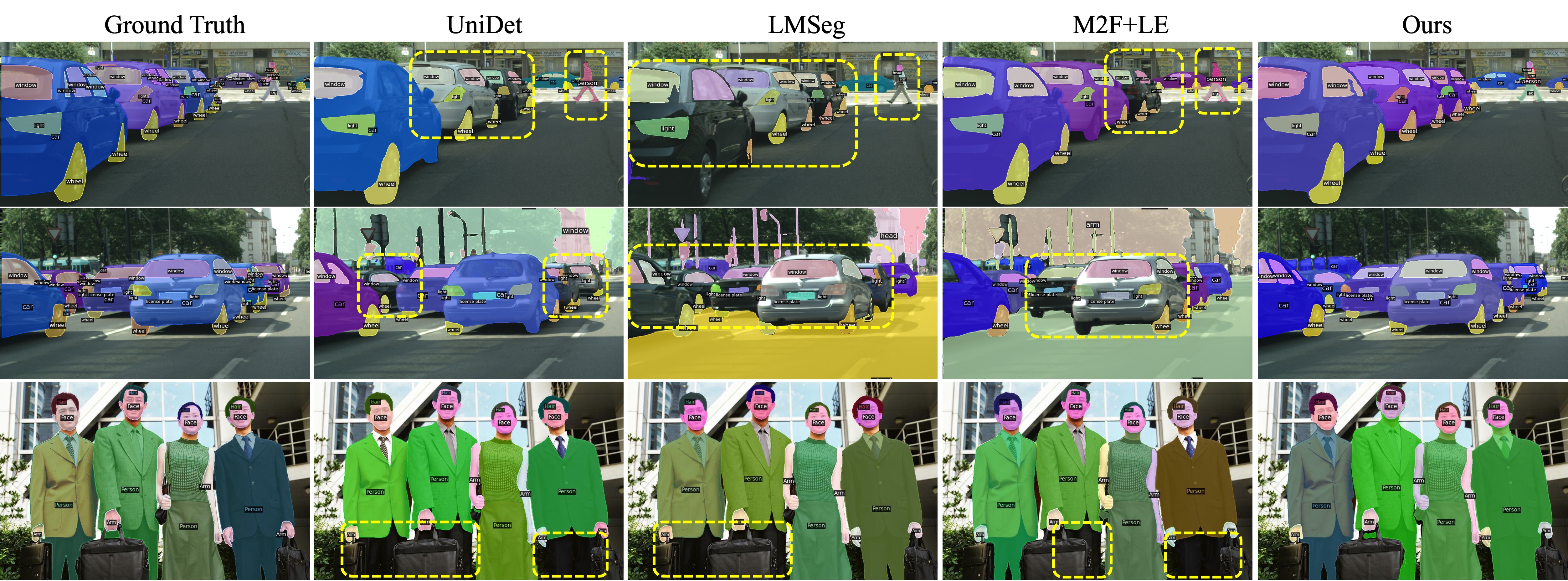}
    \caption{Visual comparison of multi-category segmentation performance. We present a overview of \ourmethodname's capabilities in handling complex label spaces. The depicted scenarios demonstrate the model's proficiency in simultaneous cross-label space multi-category predictions, a task where traditional segmentation approaches often fall short.}
    \vspace{-5pt}
    \label{fig:qualitative_results}
\end{figure*}

\begin{table*}[!htbp]
    \centering
    \footnotesize
    \captionsetup{font=normalsize}    
    \captionsetup[subtable]{font=small}
    \begin{subtable}{\textwidth}
        \centering
        \setlength{\tabcolsep}{10.8pt}
        \footnotesize
        \begin{tabular}{ll c ccc ccc ccc ccc}
            \multirow{2}{*}{Methods} & \multicolumn{3}{c}{\cihppair} & \multicolumn{3}{c}{\cihpmulti} & \multicolumn{3}{c}{\csppair} & \multicolumn{3}{c}{\cspmulti} \\
            \cmidrule(lr){2-4} \cmidrule(lr){5-7} \cmidrule(lr){8-10} \cmidrule(lr){11-13}
              & PQ & SQ & RQ   & PQ & SQ & RQ & PQ & SQ & RQ   & PQ & SQ & RQ \\
            \toprule
            UniDet~\cite{zhou_cvpr22_simple}   & 45.2 & 77.8 & 57.2 & 53.5 & 77.7 & 66.5 & 35.2 & 76.8 & 45.0 & 21.7 & 74.9 & 28.4 \\
            LMSeg~\cite{zhou2023lmseg}         & 41.7 & 77.0 & 53.3 & 52.4 & 77.4 & 65.3 & 37.6 & 77.2 & 47.8 & 22.9 & 73.8 & 30.4 \\
            M2F+LE~\cite{cheng2021mask2former} & 45.2 & 79.3 & 56.5 & 53.6 & 78.2 & 66.3 & 38.8 & 78.0 & 48.7 & 22.3 & 74.6 & 29.3 \\
            \ourmethodname (Ours)                    & \textbf{61.5} & \textbf{84.8} & \textbf{71.5} & \textbf{58.9} & \textbf{81.1} & \textbf{70.2} & \textbf{45.1} & \textbf{78.4} & \textbf{57.1} & \textbf{31.5} & \textbf{75.4} & \textbf{41.4} \\
            \bottomrule
        \end{tabular}
        \caption{Panoptic segmentation}
        \label{subtbl:first}
    \end{subtable}
    
    \begin{subtable}{0.47\textwidth}
        \centering
        \setlength{\tabcolsep}{5.8pt}
        \footnotesize
        \begin{tabular}{l ccc ccc}
            \multirow{2}{*}{Methods}  & \multicolumn{3}{c}{\cihppairinst} & \multicolumn{3}{c}{\csppairinst} \\
            \cmidrule(lr){2-4} \cmidrule(lr){5-7} 
            & $AP$ & $AP_{50}$ & $AP_{75}$ & $AP$ & $AP_{50}$ & $AP_{75}$ \\
            \toprule
            UniDet~\cite{zhou_cvpr22_simple}    & 23.7 & 45.9 & 22.1 & 18.1 & 34.0 & 17.4 \\
            LMSeg~\cite{zhou2023lmseg}          & 21.8 & 42.4 & 20.1 & 18.8 & 35.4 & 17.8 \\
            M2F+LE~\cite{cheng2021mask2former}  & 24.8 & 46.2 & 23.8 & 19.9 & 36.1 & 19.3 \\
            \ourmethodname (Ours)                     & \textbf{44.3} & \textbf{68.8} & \textbf{48.4} & \textbf{24.6} & \textbf{44.9} & \textbf{23.2} \\
            \bottomrule
        \end{tabular}
        \caption{Instance segmentation}
        \label{subtbl:second}
    \end{subtable}
    \hfill
    \begin{subtable}{0.48\textwidth}
        \centering
        \setlength{\tabcolsep}{4.5pt}
        \footnotesize
        \begin{tabular}{cccc cccc}
            \multicolumn{4}{c}{\cocosemseg} & \multicolumn{4}{c}{\cssemseg} \\
            \cmidrule(lr){1-4} \cmidrule(lr){5-8} 
            mIoU & fwIoU & mACC & pACC & mIoU & fwIoU & mACC & pACC \\
            \toprule
            58.9  & 67.9  & 71.4  & 79.7  & 75.6  & 50.8  & 89.3 & 68.4  \\
            59.2  & 68.1  & 71.7  & 79.7  & 73.6  & 50.6  & 89.2 & 68.5  \\
            58.5  & 68.0  & 71.4  & 79.7  & 75.4  & 52.6  & 89.2 & 67.6  \\
            \textbf{59.4}  & \textbf{68.2}  & \textbf{72.0}  & \textbf{79.9}  & \textbf{78.2}  & \textbf{62.3}  & \textbf{90.4} & \textbf{80.9}  \\
            \bottomrule
        \end{tabular}
        \caption{Semantic segmentation}
        \label{subtbl:third}
    \end{subtable}
    \caption{Performance improvements on various datasets across three segmentation tasks: panoptic, instance, and semantic.}
    \label{tbl:results_3tables_merged}
\end{table*}

\setlength{\tabcolsep}{4.2pt}
\renewcommand{\arraystretch}{1.0}
\begin{table}[!htbp]\begin{center}
{
\footnotesize
\begin{tabular}{l cccccc}
Methods & $PIQ$ & $PIQ_{50}$ & $PIQ_{75}$ & $PIQ_{s}$ & $PIQ_{m}$ & $PIQ_{l}$  \\
\toprule
UniDet~\cite{zhou_cvpr22_simple}    & 41.8 & 49.8 & 41.5 & 36.8 & 48.8 & 54.4  \\
LMSeg~\cite{zhou2023lmseg}          & 42.4 & 50.7 & 41.9 & 36.9 & 49.7 & 52.1  \\
M2F+LE~\cite{cheng2021mask2former}  & 42.9 & 51.1 & 42.6 & 36.8 & 50.5 & 57.7  \\
\ourmethodname (Ours)                     & \textbf{45.9} & \textbf{56.0} & \textbf{45.2} & \textbf{39.2} & \textbf{54.2} & \textbf{60.8} \\
\bottomrule
\end{tabular}
}
\caption{Comparative analysis of Panoptic Instance Quality (PIQ) on the Cityscapes Panoptic Parts benchmark with overlapping masks for thing categories.}
\label{table:results_overlap_pano_seg}
\vspace{-2em}
\end{center}
\end{table}
\setlength{\tabcolsep}{1.4pt}
\renewcommand{\arraystretch}{1}

\setlength{\tabcolsep}{3.2pt}
\renewcommand{\arraystretch}{1.0}
\begin{table}[!htbp]
\begin{center}
\footnotesize
\begin{tabular}{l cc cc ccc | c}
Methods
  & COCO & CIHP
  & CS   & CSP
  & COCO
  & ADE
  & VST
  & Avg
  \\
\toprule
UniDet~\cite{zhou_cvpr22_simple}
  & 48.8 & 61.9 & 57.0 & 19.8 & 47.7 & 41.3 & 35.1 & 44.5 \\
LMSeg~\cite{zhou2023lmseg}
  & 48.3 & 61.3 & 56.8 & 23.8 & 47.4 & 40.8 & 34.2 & 44.7 \\
M2F+LE~\cite{cheng2021mask2former}
  & 48.7 & \textbf{62.0} & 57.9 & 24.7 & 47.9 & 41.3 & 33.9 & 45.2 \\
\ourmethodname (Ours)
  & \textbf{49.0} & 61.8 & \textbf{61.0} & \textbf{32.7} & \textbf{48.0} & \textbf{42.5} & \textbf{35.4} & \textbf{47.2} \\

\bottomrule
\end{tabular}
\caption{Per-dataset label space evaluation for three dataset groups (\dgrponetxt, \dgrptwotxt, \dgrpthreetxt). Each method (rows) was trained for each of the three groups and then evaluated on the individual label spaces.}
\label{table:results_labelspace_separate}
\vspace{-2em}
\end{center}
\end{table}
\setlength{\tabcolsep}{1.4pt}
\renewcommand{\arraystretch}{1}

\begin{table}[ht]
    \centering
    \footnotesize
    \setlength{\tabcolsep}{5pt}
    \begin{tabular}{cc | cccc | c}
        \rotatebox[origin=l]{0}{LSQE} & 
        \rotatebox[origin=l]{0}{\panoinffinal} &
        \rotatebox[origin=l]{0}{\cihppair} &
        \rotatebox[origin=l]{0}{\cihpmulti} &
        \rotatebox[origin=l]{0}{\csppair} &
        \rotatebox[origin=l]{0}{\cspmulti} &
        \rotatebox[origin=l]{0}{Avg} \\
    \toprule
    \xmark  & \xmark & 45.2 & 53.6 & 38.7 & 22.3 & 40.0 \\
    \cmark & \xmark & 60.2 & 56.0 & 32.9 & 11.9 & 40.3 \\
    \xmark & \cmark & 45.8 & 53.4 & 38.0 & 22.9 & 40.0 \\
    \cmark & \cmark & \textbf{61.5} & \textbf{58.9} & \textbf{45.1} & \textbf{31.5} &  \textbf{49.3} \\
    \bottomrule
    \end{tabular}
    \caption{Ablation study of different model components in \ourmethodname on average PQ. LSQE enables the model to generate diverse masks from various input categories, even in the presence of potential semantic conflicts. However, this diversity may sometimes result in a performance drop due to confusion and overlap, as seen in the second row for \csppair and \cspmulti. ESF-OMI effectively refines the masks produced by LSQE, leading to superior performance when both modules operate in tandem (last row).}
    \label{table:RESI_ablation}
    \vspace{-1em}
    \setlength{\tabcolsep}{1.4pt}
\end{table}

\begin{figure}[!htbp]
    \centering
    \includegraphics[trim={0 0.6cm 0 0},clip, width=\columnwidth]{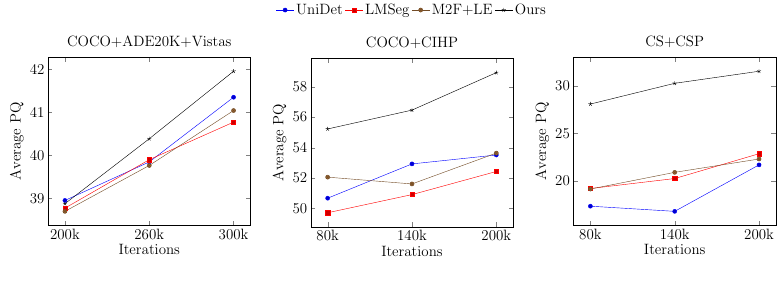}
    \caption{
    Average PQ w.r.t. total training cost for three panoptic segmentation dataset(COCO, ADE20K, Mapillary Vistas) and two mixed-label space evaluation-only datasets CIHP/CSP$_{\textrm{multi}}$.}
    \label{fig:avg_multi_PQ_training}
    \vspace{-1.5em}
\end{figure}

\paragraph{Panoptic Segmentation.} 
Tab.~\ref{tbl:results_3tables_merged} demonstrates our results on the mixed-label space benchmarks \cihppair, \cihpmulti, \csppair and \cspmulti. For each method, we evaluate two models trained on dataset groups \dgrptwotxt and \dgrpthreetxt, respectively. Our method \ourmethodname outperforms all baselines on all four benchmarks.

To ensure that our superior performance was not merely a result of differences in convergence rates or training iterations among the models, we standardized the hyperparameter settings across all models during training. This standardization allowed for a fair assessment of each model's intrinsic capabilities. As illustrated in Fig.~\ref{fig:avg_multi_PQ_training}, \ourmethodname consistently outperforms the competition across the board. The Average PQ scores, plotted with respect to training steps for three multi-dataset training settings, indicate that our model maintains a higher performance level throughout the training process.

The baselines struggle with handling the mixed label spaces and the semantic inconsistencies in the combined training datasets. To better illustrate this large performance gap, we visualize qualitative results in Fig.~\ref{fig:qualitative_results}. These results clearly show imperfect mask predictions for the baseline, indicating a significant limitation in their ability to adapt to complex segmentation scenarios.

In addition to these observations, we also noted a strong generalization ability of our model across diverse paired category situations (results provided in supplementary materials). This is demonstrated in a qualitative comparison in handling user-specified, random category combinations. We see \ourmethodname consistently delivers accurate segmentations for paired random categorical combinations chosen across all label spaces from different datasets. 

\paragraph{Panoptic post-processing.}
\label{para:pano_post_processing}

Models influenced by Mask2Former, such as \ourmethodname, generate a set of $N$ masks (or $N * |D|$ in the case of \ourmethodname), each accompanied by a probability distribution across the label space. The decoder's self-attention layers help correlate predictions; however, overlapping masks can still occur. To produce a coherent and non-overlapping segmentation output, post-processing is essential to resolve overlaps and create a unified segmentation mask that accurately represents different objects and regions. The original Mask2Former method does not adequately address the challenge of overlapping but accurate masks, especially in multi-dataset scenarios. Additionally, when smaller objects (like sunglasses on a person) overlap with larger ones, the original algorithm often fails to retain these smaller objects. To improve this, we introduce ESF-OMI, which refines how overlapping masks are handled. Surprisingly, simple modifications to the existing inference algorithm, as done in ESF-OMI, effectively resolve the issue and significantly improve segmentation accuracy. More details and pseudocode of ESF-OMI can be found in the supplementary materials.

We next investigate the impact of different model components. We evaluate ESF-OMI on the proposed model \ourmethodname and one baseline, M2F+LE. As shown in Tab.~\ref{table:RESI_ablation}, the proposed inference algorithm \panoinffinal is crucial when handling semantically overlapping label spaces. For non-overlapping spaces, both methods perform similarly, with the original algorithm~\cite{cheng2021maskformer} having a slight advantage. However, the post-processing algorithm alone does not account for the entire performance gap between \ourmethodname and the baselines in Tab.~\ref{subtbl:first}. Even with the original post-processing, \ourmethodname outperforms the baselines. Additionally, we assess per-dataset label-space performance in Tab.~\ref{table:results_labelspace_separate}, demonstrating LSQE's effectiveness in resolving semantic inconsistencies during multi-dataset training.

\paragraph{Instance Segmentation.} Furthermore, we also investigate whether the baselines struggle only with the post-processing (all methods are based on Mask2Former~\cite{cheng2021mask2former}) or with the mask prediction in the first place. To do so, we evaluate the models on instance segmentation which does not require any post-processing as it allows overlapping masks. Tab.~\ref{subtbl:second} demonstrates that all baselines struggle already in predicting correct masks, which matches our observations from the qualitative results in Fig.~\ref{fig:qualitative_results}.

\paragraph{Semantic Segmentation.} Additionally, we extended our evaluation to semantic segmentation tasks to further assess the versatility of our model. Tab.~\ref{subtbl:third} presents these results, where \ourmethodname outperforms all baselines. This superior performance in semantic segmentation, a task focusing on per-pixel classification without the complexity of instance delineation, offers a purer assessment of \ourmethodname's capability in discerning and categorizing diverse label spaces.

\paragraph{Panoptic Instance Segmentation.}
In addition to panoptic, instance, and semantic segmentation, we evaluate the Panoptic Instance Quality (PIQ) on the Cityscapes Panoptic Parts benchmark, which features overlapping masks for ``thing'' categories. As detailed in Sec. \ref{sec:benchmarkingSeg} \emph{Benchmarking Image Segmentation}, our aim is to combine the benefits of per-pixel classification from panoptic segmentation with the allowance for overlapping masks in instance segmentation, thereby enhancing the accuracy measurement of a model in a unified and scalable manner. Tab.~\ref{table:results_overlap_pano_seg} demonstrates that \ourmethodname outperforms all baselines.

\subsection{Evaluation on Per-dataset Label Spaces}
\label{sec:4_3}

Next, we evaluate all models on the label spaces of the individual datasets for each of the three training dataset groups, \dgrponetxt, \dgrptwotxt, \dgrpthreetxt. As shown in Tab.~\ref{table:results_labelspace_separate}, all four models perform similarly well across most benchmarks. This demonstrates that our adapted inference algorithm, which runs the decoder multiple times (Sec.~\ref{sec:method_partwhole_relations}), remains effective in this setting while clearly outperforming the baselines in the mixed-label space setting (Tab.~\ref{subtbl:first}). Notably, even during per-dataset label space evaluation, the decoder in \ourmethodname runs multiple times if two of the training datasets share the same category. We focus on PQ in this setting to facilitate comparison with prior works that only conducted per-dataset evaluations.

One standout result in Tab.~\ref{table:results_labelspace_separate} is the significantly higher PQ of \ourmethodname on the benchmarks CS and CSP. The reason is that in dataset group \dgrpthreetxt, the training images are exactly the same but the annotations are different (whole objects versus parts in CityScapes~\cite{Cordts2016Cityscapes}). This seems to confuse all baseline models while \ourmethodname can handle this. Fig.~\ref{fig:qualitative_results} shows some examples. Note that this is even a practical use case where an existing annotated dataset is extended with new labels but not for all images to save cost.

\begin{table}[ht]
    \centering
    \footnotesize
    \setlength{\tabcolsep}{3pt}
    \begin{tabular}{l|c|c}
    Methods & \shortstack[c]{Max GFlops: ($10^9$ ops/s)} & \shortstack[c]{Avg Time: (s/iter/device)} \\
        \toprule
        UniDet~\cite{zhou_cvpr22_simple}      & 272.4 ± 0.9    & 0.1530 \\ 
        LMSeg~\cite{zhou2023lmseg}       & 273.0 ± 0.9    & 0.1655 \\ 
        M2F+LE~\cite{cheng2021mask2former}     & 272.5 ± 0.9    & 0.1593 \\ 
        \ourmethodname 1 dataset  & 272.5 ± 0.9    & 0.1593 \\ 
        \ourmethodname 2 datasets & 360.1 ± 0.9    & 0.2756 \\ 
        \ourmethodname 3 datasets & 447.7 ± 0.9    & 0.3919 \\ 
     \bottomrule
    \end{tabular}
    \caption{Max total GFlops and pure compute time for all methods (average of 500 inferences, batch size 16, and 100 queries).}
    \label{table:time_rebuttal}
    \setlength{\tabcolsep}{1.4pt}
    \vspace{-1.5em}
\end{table}

\section{Conclusion}
\label{sec:conclusions}

Leveraging multiple existing datasets to train image segmentation models is a cost-effective way to scale up and is crucial for improving robustness and semantic understanding. However, multi-dataset training becomes challenging when mixing label spaces leads to inconsistent semantics. While prior methods struggle, our proposed model, \ourmethodname, directly addresses these inconsistencies with learnable label space-specific parameters and novel inference strategies. Extensive experiments show that \ourmethodname effectively handles complex label spaces, with a negligible impact on model size and a slight increase in inference time. For future work, we plan to explore more efficient methods to merge label spaces and resolve conflicts in an open-vocabulary setting.

{\small
\bibliographystyle{ieee_fullname}
\bibliography{egbib}
}

\clearpage

\maketitlesupplementary
\appendix
\noindent

In this supplementary material, we provide additional details and results that were not included in the main paper due to space constraints. In Sec. \ref{supp_sec:post_processing}, we give pseudo code and details of our post-processing strategy for panoptic segmentation. In Sec. \ref{supp_sec:benchmarks}, we provide details of how we construct the testing benchmarks. In Sec. \ref{supp_sec:qualitative}, we give more qualitative comparisons. Finally, in Sec. \ref{supp_sec:training_details}, we introduce the details of our model training.

\section{Panoptic Segmentation Post-processing}
\label{supp_sec:post_processing}

\begin{figure*}[!htbp]\centering
    \includegraphics[width=\textwidth]{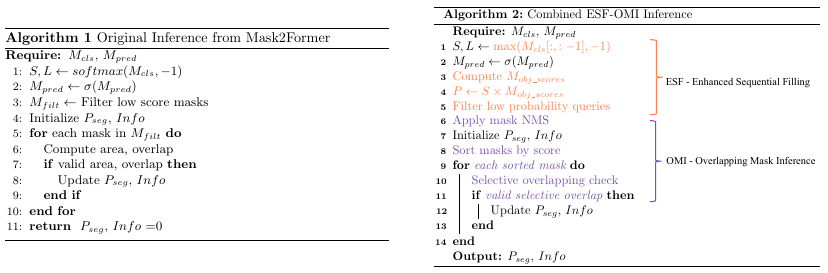}
    \caption{Pseudo code of the original post-processing algorithm from~\cite{cheng2021maskformer} and our proposed algorithm, ESF-OMI, which aims to resolve overlapping mask predictions for panoptic segmentation.}
    \label{fig:pseudo}
\end{figure*}

\begin{figure}[!htbp]\centering
    \includegraphics[width=1\columnwidth]{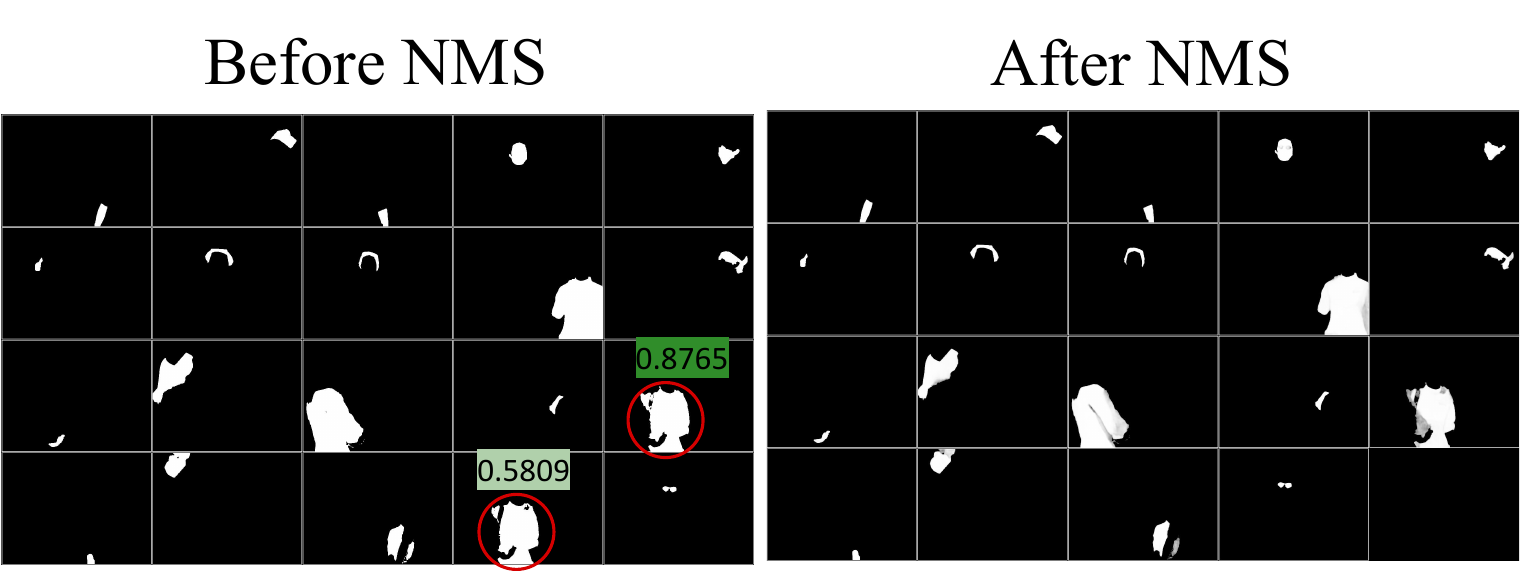}
    \caption{An illustration of the mask-NMS used in the proposed post-processing algorithm, ESF-OMI. The figure shows how near-duplicate masks are removed in panoptic segmentation.}
    \label{fig:example_nms}
\end{figure}

\begin{figure}[!htbp]\centering
    \includegraphics[width=1\columnwidth]{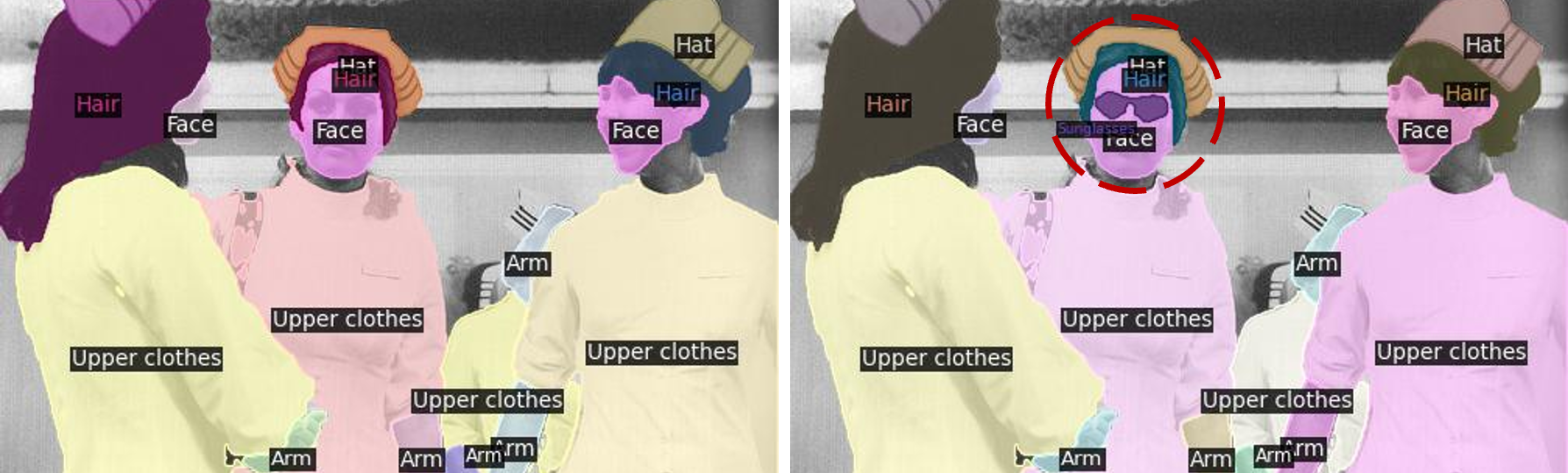}
    \caption{Example output of the original panoptic post-processing algorithm from \cite{cheng2021maskformer} (left) versus the output of the proposed post-processing ESF-OMI (right). For panoptic segmentation task, our proposed algorithm better handles overlapping mask predictions like the sunglasses in the figure (red circle), which are suppressed by the original algorithm.
    }
    \label{fig:algorithm_vis}
\end{figure}

Both algorithms are summarized with pseudo code in Fig.~\ref{fig:pseudo}, and both receive as input mask predictions $M_{pred}$ with class probabilities $M_{cls}$ and output the panoptic segmentation map $P_{seg}$. The original algorithm first computes the most likely class ($L$) and confidence ($S$) for each mask, and filters out low-scoring ones or those that are assigned the background class. The panoptic map $P_{seg}$ is then iteratively filled, starting with the most confident mask. A new mask is only added if it occupies an appropriate area of the image and is not too small (``if vavlid area, overlap''). The proposed method ESF-OMI makes two key adjustments, which are highlighted in orange and purple in Fig.~\ref{fig:pseudo}.
(1) Masks are filtered in a different way. The background class is excluded from the filtering step and all masks with a score above a threshold survive the filtering. Note there can be class confidences of 0.3 for one class, but 0.7 for background -- this mask is filtered in the original algorithm but kept in ESF-OMI if the threshold is below 0.3. (2) The criteria for placing masks on the pantopic segmentation map $P_{seg}$ are different. First, a non-maxima-suppression (NMS) step based on masks removes near-duplicates, which would otherwise lead to noisy outputs, see Fig.~\ref{fig:example_nms}. Second, when placing masks on the segmentation map the criterion ``valid selective overlap'' allows smaller masks $M_S$ (with a lower score) to be placed on top of an existing mask $M_B$ in $P_{seg}$ if $M_S$ is fully contained in $M_B$ (with some slack). This ensures that smaller objects are not omitted in the final segmentation, like sunglasses on a person as illustrated in Fig.~\ref{fig:algorithm_vis}.

\section{Mixed-label Space Benchmarks}
\label{supp_sec:benchmarks}
As stated in the main paper, we build multiple evaluation-only mixed-label space benchmarks to properly evaluate the ability of multi-dataset models to handle any combination of label spaces $A$ and $B$ of the individual datasets. Label space $C$ must contain two partitions that include categories exclusively from either training dataset, i.e., $| C \cap (A \setminus B) | > \emptyset \land |C \cap (B \setminus A)| > \emptyset$. This can be extended to more than two training datasets easily.

\begin{figure}[!htbp]
\centering
    \includegraphics[width=0.98\linewidth]{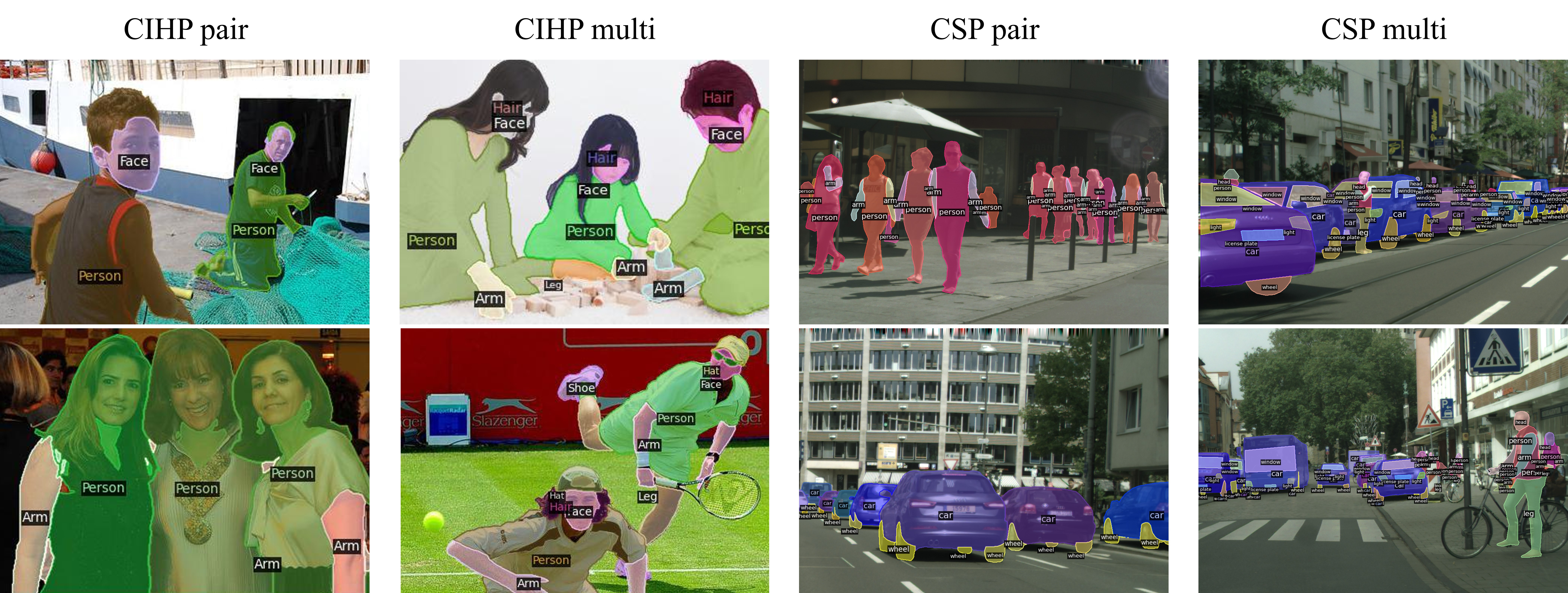}
    \caption{More examples of the mixed-label space evaluation-only datasets: CIHP$_{\textrm{P}}$, CIHP$_{\textrm{M}}$, CSP$_{\textrm{P}}$, and CSP$_{\textrm{M}}$.}
    \label{fig:benchmark_vis}
\end{figure}

Here, we show all the sub-dataset label spaces in each of the mixed-label space benchmarks (more visualizations can be found in Fig.~\ref{fig:benchmark_vis}):

\begin{itemize}
    \item \supcihpp: [arm, person], [coat, person], [dress, person], [face, person], [glove, person], [hair, person], [hat, person], [leg, person], [pants, person], [scarf, person], [shoe, person], [skirt, person], [socks, person], [sunglasses, person], [upper clothes, person]
    \item \supcihpm: [leg, shoe, person], [hat, hair, face, person], [hat, hair, face, arm, leg, person]
    \item \supcspp: [window, car], [wheel, car], [light, car], [license plate, car], [head, person], [arm, person], [leg, person]
    \item \supcspm: [license plate, light, wheel, window, car, arm, head, leg, person]
\end{itemize}

\begin{figure}[!htbp]
\centering
    \includegraphics[width=1\linewidth]{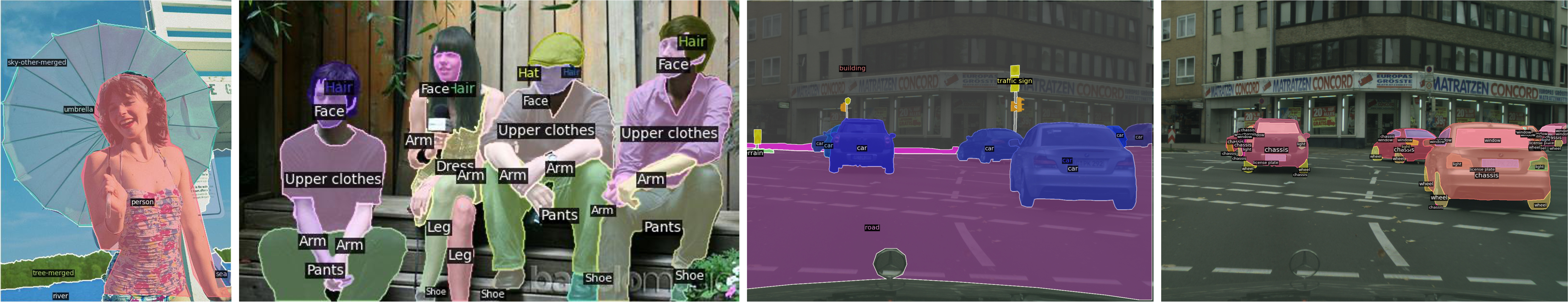}
    \caption{Examples of the ground truth annotations for the original training datasets (from left to right): COCO, CIHP, CS, and CSP.}
    \label{fig:training_gt_vis}
\end{figure}

The ground truth annotations for the original training datasets are shown in Fig.~\ref{fig:training_gt_vis} for reference.

\section{Additional Qualitative Comparison}
\label{supp_sec:qualitative}
\begin{figure*}[!htbp]\centering
    \includegraphics[width=\textwidth]{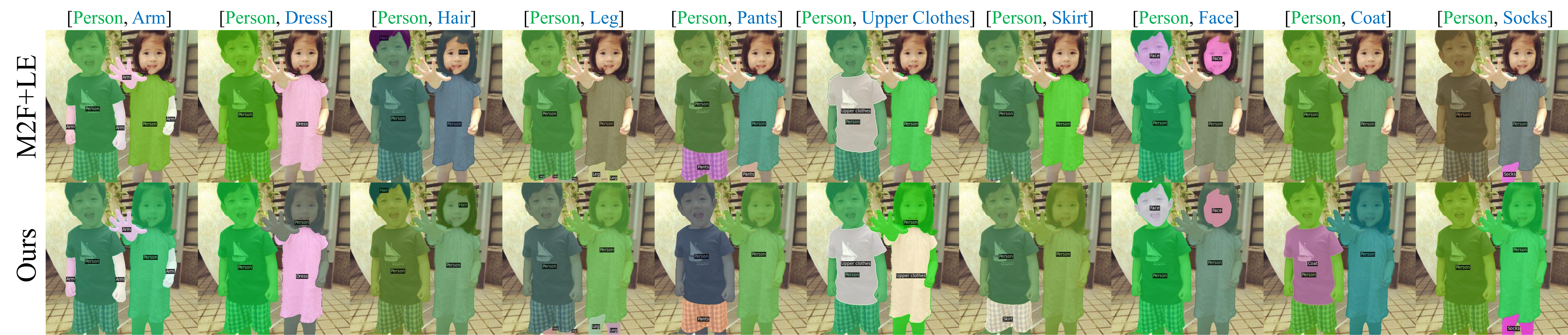}
    \caption{
    More visual comparisons of pair-category segmentation performance. \ourmethodname consistently demonstrates superior results in handling diverse category combinations during inference compared to conventional approaches, exemplified here by Mask2Former+LE~\cite{cheng2021mask2former}.}
    \label{fig:more_pair_vis}
\end{figure*}

\begin{figure*}[!htbp]\centering
    \includegraphics[width=1\textwidth]{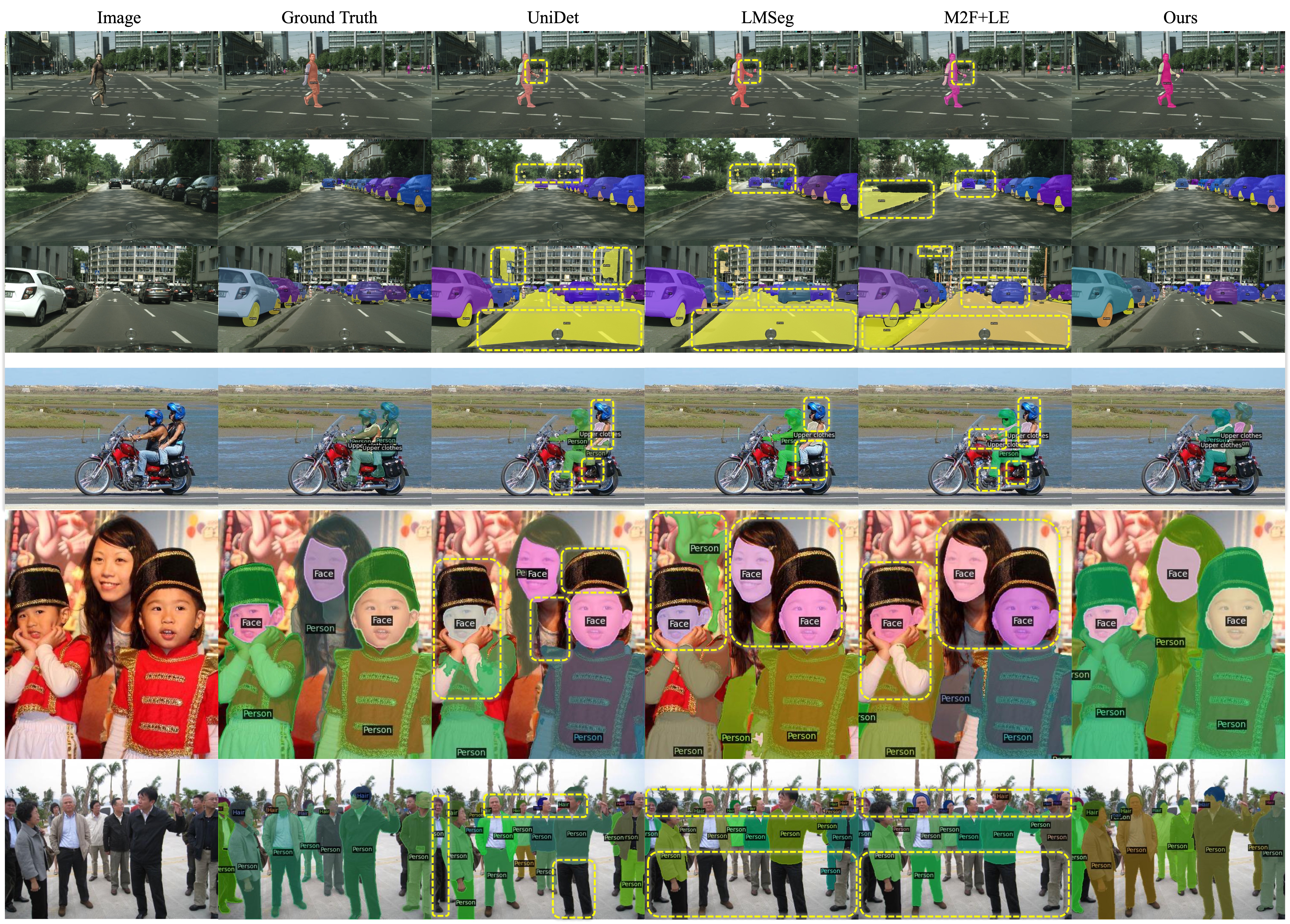}
    \caption{More visual comparison of pair-category segmentation performance across all models on \cihppair and \csppair}
    \label{fig:pair_vis_full}
\end{figure*}

\begin{figure*}[!htbp]\centering
    \includegraphics[width=\textwidth]{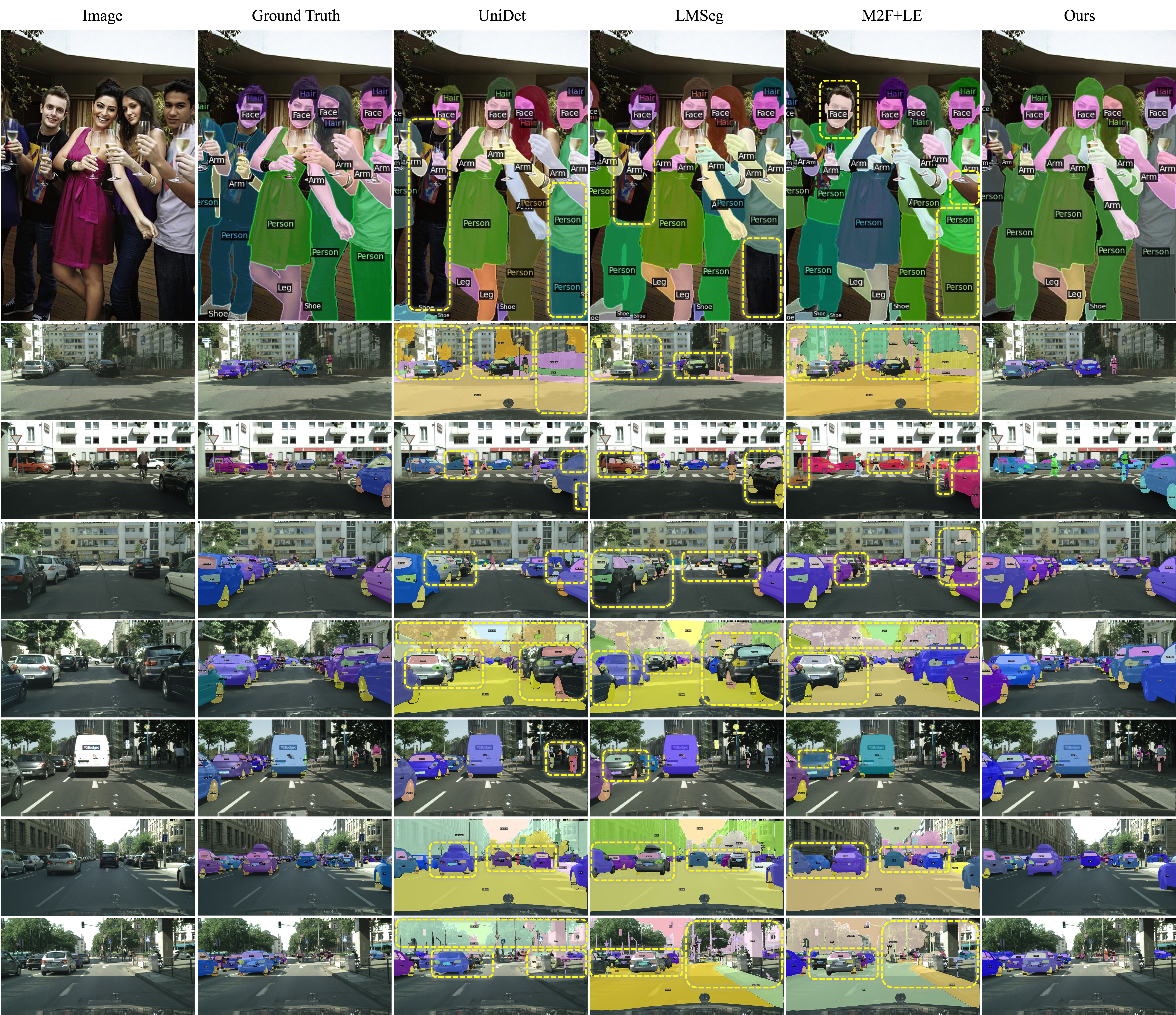}
    \caption{More visual comparison of multi-category segmentation performance across all models on \cihpmulti and \cspmulti.}
    \label{fig:multi_vis_full}
\end{figure*}

We provide a visual comparison to showcase the qualitative performance of each model, alongside the original image and ground truth annotations.
\begin{enumerate}
    \item \ourmethodname excels in handling complex class combinations (e.g., ``license plate, light, wheel, window, car, arm, head, leg, person''). See Fig.~\ref{fig:multi_vis_full}.
    \item \ourmethodname also shows versatility across various class combinations, outperforming others in Fig.~\ref{fig:pair_vis_full} and Fig.~\ref{fig:more_pair_vis}.
\end{enumerate}

As stated in Sec. \textcolor{red}{3.4} - Naive Approach, we observed that even when including language-based embeddings as classifiers for existing instance segmentation models, the resulting models often struggle with semantically inconsistent relationships between label spaces. Examples are shown in the first row of Fig.~\ref{fig:more_pair_vis}.

\section{Model Training Details}
\label{supp_sec:training_details}
We use ResNet-50 (R50) as our backbone across all experiments. For each multi-dataset training method, we train the model for 200k iterations on COCO-CIHP and Cityscapes-CPP, and 300k iterations on COCO, ADE20K, and Mapillary Vistas. We use a batch size of 16 and train on 8 A100 GPUs. To accommodate the different sizes of the multiple datasets employed, we implement a data sampling scheme that aims to sample images from each dataset with equal frequency, as described in UniDet~\cite{zhou_cvpr22_simple}. Information regarding the evaluation dataset setup can be found in Mixed-label Space Benchmarks(Sec.~\ref{supp_sec:benchmarks}). Further details about the panoptic inference algorithm setup can be found in Panoptic Segmentation Post-processing(Sec.~\ref{supp_sec:post_processing}).

\end{document}